\documentclass{article}

\usepackage{PRIMEarxiv}

\usepackage[utf8]{inputenc} % allow utf-8 input
\usepackage[T1]{fontenc}    % use 8-bit T1 fonts
\usepackage{hyperref}       % hyperlinks
\usepackage{url}            % simple URL typesetting
\usepackage{booktabs}       % professional-quality tables
\usepackage{amsfonts}       % blackboard math symbols
\usepackage{nicefrac}       % compact symbols for 1/2, etc.
\usepackage{microtype}      % microtypography
\usepackage{lipsum}
\usepackage{fancyhdr}       % header
\usepackage{graphicx}       % graphics
\graphicspath{{media/}}     % organize your images and other figures under media/ folder
\usepackage{algorithm}      % For the algorithm environment
\usepackage{algpseudocode}  % For the pseudocode environment
\usepackage{float}          % For precise float placement with [H]
\usepackage{tikz}
\usepackage{tikz-3dplot}
\usepackage{xcolor}
\usepackage{url}
\usepackage{siunitx}
\usepackage{multirow}
\usepackage{natbib}
\usepackage{amsmath}

\title{Curriculum-Based Iterative Self-Play for Scalable Multi-Drone Racing
\thanks{%\textit{\underline{Citation}}: 
\textbf{This paper is currently under review at the journal Engineering Applications of Artificial Intelligence.}} 
}

\author{
  Onur Akgün \\
  Department of Mechatronics Engineering \\
  Faculty of Engineering \\
  Turkish-German University \\
  Istanbul, Türkiye\\
  \texttt{akgun@tau.edu.tr} \\
}

\begin{document}
\maketitle

\begin{abstract}
The coordination of multiple autonomous agents in high-speed, competitive environments represents a significant engineering challenge. This paper presents CRUISE (Curriculum-Based Iterative Self-Play for Scalable Multi-Drone Racing), a reinforcement learning framework designed to solve this challenge in the demanding domain of multi-drone racing. CRUISE overcomes key scalability limitations by synergistically combining a progressive difficulty curriculum with an efficient self-play mechanism to foster robust competitive behaviors. Validated in high-fidelity simulation with realistic quadrotor dynamics, the resulting policies significantly outperform both a standard reinforcement learning baseline and a state-of-the-art game-theoretic planner. CRUISE achieves nearly double the planner’s mean racing speed, maintains high success rates, and demonstrates robust scalability as agent density increases. Ablation studies confirm that the curriculum structure is the critical component for this performance leap. By providing a scalable and effective training methodology, CRUISE advances the development of autonomous systems for dynamic, competitive tasks and serves as a blueprint for future real-world deployment.
\end{abstract}

% keywords can be removed
\keywords{autonomous drone racing \and self-play \and curriculum learning \and deep reinforcement learning \and multi-agent systems \and adaptive racing strategies}

\section{Introduction}
\label{sec:introduction}
The effective coordination of multiple autonomous agents is a fundamental challenge in modern engineering, with critical applications in logistics automation, autonomous transportation, and environmental monitoring. These real-world scenarios often involve dynamic, competitive, or collaborative elements where agents must make split-second decisions in cluttered environments. Among these challenges, high-speed, competitive multi-agent racing serves as a particularly demanding benchmark, pushing the limits of agility, safety, and scalable coordination \citet{moon2019challenges, foehn2022alphapilot}. While reinforcement learning (RL) has shown champion-level performance for single agents \citet{kaufmann2023champion}, scaling these successes to the multi-agent domain remains a significant hurdle due to issues of non-stationarity, computational complexity, and safe exploration \citet{wang2020multi}.

Developing and testing control policies for such high-stakes physical systems presents significant practical barriers. The process involves prohibitive hardware costs, substantial safety risks to equipment and personnel, and the difficulty of creating perfectly reproducible test conditions. Consequently, high-fidelity simulation has become an essential and standard tool in the engineering design life-cycle \citet{guerra2019flightgoggles, song2021flightmare}. It enables the initial design, rigorous validation, and comprehensive benchmarking of complex multi-agent control policies before real-world deployment. However, even within simulation, training robust and scalable policies efficiently remains an open problem, necessitating novel frameworks that can manage complexity and reduce sample requirements.

To address this gap, this paper introduces CRUISE (Curriculum-Based Iterative Self-Play for Scalable Multi-Drone Racing), a novel training framework designed to produce high-performance, scalable policies for complex multi-agent coordination tasks. CRUISE synergistically integrates two key principles: (1) a progressive curriculum that simplifies the learning problem by gradually increasing the number of opponents and task difficulty, and (2) an efficient iterative self-play mechanism that fosters robust competitive behaviors by training against an evolving set of frozen, high-performing opponent policies. This combined approach is specifically designed to enhance sample efficiency and ensure scalability—two of the most critical factors for complex robotics applications.

We validate the CRUISE framework within the challenging domain of multi-drone racing, using a high-fidelity simulator with realistic quadrotor dynamics. \footnote{A supplementary video detailing our method and demonstrating its performance is available at \\ \url{https://drive.google.com/file/d/1k7necen2DgIxaYT2alKK8-b20sE_AyDA/view}}. Our results show that CRUISE policies substantially outperform both a standard RL baseline and a state-of-the-art game-theoretic planner in terms of racing speed, success rate, and collision avoidance (Sec. ~\ref{sec:results}). The contributions of this work are therefore threefold. First, we propose a novel and generalizable training framework, CRUISE, that integrates curriculum learning with iterative self-play for efficient and scalable multi-agent coordination. Second, we provide comprehensive empirical validation of the framework, including ablation studies that confirm the critical role of the curriculum structure in achieving high performance. Finally, we support open and reproducible science by providing the full open-source release of our code, environments, and pretrained models \footnote{Full implementation details, including all hyperparameters, are available at \\ \url{https://doi.org/10.5281/zenodo.17256943}}

The remainder of this paper details the CRUISE methodology (Sec.~\ref{sec:methodology}), presents the empirical results (Sec.~\ref{sec:results}), discusses their implications (Sec.~\ref{sec:discussion}), and offers a conclusion and directions for future work (Sec.~\ref{sec:conclusion}, Sec.~\ref{sec:future_work}).

\section{Related Works}
\label{sec:related_work}

Autonomous multi-drone racing research spans both planning-based and learning-based methods. Game-theoretic planners \citet{spica2020real, di2023cooperative} have been developed to address the competitive aspects of racing. Notably, Sensitivity Enhanced Iterative Best Response (SE-IBR) \citet{wang2020multi}, which extends game theory to 3D racing, serves as a key state-of-the-art planning baseline in our work. However, these planners often face scalability challenges with an increasing number of agents and rely on accurate dynamics models. In parallel, Reinforcement Learning (RL) has excelled in single-agent drone racing, with policies achieving champion-level performance against human pilots \citet{kaufmann2018deep, loquercio2019deep, kaufmann2023champion} and demonstrating competitiveness with optimal control \citet{song2023reaching}. However, applying Multi-Agent RL (MARL) directly to this domain introduces significant scalability and sample efficiency hurdles due to the fast, adversarial physical interactions. Standard RL frameworks (e.g., Stable-Baselines3 \citet{stable-baselines3}, used for our VANILLA baseline) typically struggle in MARL without substantial enhancements.

To tackle these challenges, the community has explored innovations across multiple axes. Concurrent work has focused on achieving time-optimal, aggressive flight through sophisticated reward engineering, demonstrating impressive sim-to-real transfer by training the full multi-agent system directly \citet{wang2025dashing}. Other advances have improved agent capabilities by enhancing sample efficiency through the design of information-dense, structured observations for navigation \citet{xu2025sigmarl}, or by developing novel communication architectures like the Graph Diffusion Network to improve coordination in cooperative tasks \citet{jiang2025graph}. A distinct paradigm is Evolutionary Robotics, which uses Evolutionary Algorithms to optimize the parameters of a pre-defined, often simpler, controller rather than learning a complex policy from scratch, a technique demonstrated effectively in real-world swarm formations \citet{stolfi2024evolutionary}. While powerful, these approaches focus on optimizing agent perception, communication, or the parameters of a fixed control law, rather than the learning process for emergent strategy.

Perhaps the most relevant direction to our work is the use of curriculum learning (CL) and self-play (SP), which have proven effective in domains ranging from simulated sports \cite{lin2023tizero} to strategy games \cite{silver2017mastering, le2025pommerman}. In robotics, curriculum-based approaches have been diverse. For cooperative tasks, curricula have been implemented via staged reward functions that guide agents through pre-defined sub-tasks like formation and tracking \cite{wang2025autonomous}. This concept has been extended further by designing complex, theoretically-grounded rewards based on classical flocking models, which implicitly encode a curriculum by guiding agents toward a provably stable equilibrium \cite{guo2025invulnerable}. Other curricula have focused on improving generalization for a single agent, either by creating versatile, generalist policies via Multi-Task RL \cite{xing2024multi} or by automatically generating novel race tracks \cite{xing2025environment} or enabling diverse missions without retraining through goal-conditioning \citet{kim2024fully}. Sample efficiency has also been a direct target, with methods that strategically re-sample challenging initial states to focus the learning process \cite{messikommer2024contrastive}.

The CRUISE framework is positioned distinctly from these diverse approaches. It addresses the unique challenge of scalable learning for emergent competitive strategy. Our work investigates whether complex, adversarial behaviors like blocking and overtaking can be learned end-to-end, without the aid of explicit hierarchical planners, heavily engineered reward curricula \cite{wang2025autonomous}, or pre-defined control laws \cite{stolfi2024evolutionary}. Crucially, unlike methods that encode the solution into a complex, theoretically-grounded reward function \cite{guo2025invulnerable}, CRUISE utilizes a simple, sparse reward and instead engineers the learning process itself. We demonstrate that a structured curriculum, implemented through iterative self-play, provides a powerful and scalable methodology to solve the severe exploration problem in head-to-head competition, fostering the emergence of sophisticated tactics. This focus on scalable learning within a competitive multi-agent context is a distinct and complementary contribution to the goal of single-agent task generalization \cite{xing2024multi}. This positions our work as a contribution to understanding how to structure the learning process to enable scalable, emergent strategy in complex, competitive multi-agent domains.

\section{Methodology}
\label{sec:methodology}
This section details the CRUISE framework, beginning with the drone modeling and control architecture before describing our core contribution: a reinforcement learning approach integrating a structured curriculum with iterative self-play.

\subsection{Drone Dynamics and Controller Design}
We model the drone using standard quadrotor dynamics and employ a hierarchical controller that allows the RL policy to operate at a high level of abstraction.

\subsubsection{Notation and Dynamics}
\begin{figure}[t]
    \centering
    \begin{tikzpicture}[scale=1.2]
        % Set viewing angle
        \tdplotsetmaincoords{60}{120}
        
        \begin{scope}[tdplot_main_coords]
            % World frame - positioned clearly at bottom left
            \coordinate (W) at (-5,-7,-5);
            \draw[-stealth,red,very thick] (W) -- ++(1,0,0) node[anchor=west]{$x_W$};
            \draw[-stealth,green!70!black,very thick] (W) -- ++(0,1,0) node[anchor=south]{$y_W$};
            \draw[-stealth,blue,very thick] (W) -- ++(0,0,1) node[anchor=south west]{$z_W$};
            
            % Quadrotor body - "+" configuration
            \coordinate (B) at (0,0,0);
            
            % Arms
            \draw[thick] (-2.5,0,0) -- (2.5,0,0);
            \draw[thick] (0,-2.5,0) -- (0,2.5,0);
            
            % Rotors with rotation direction and thrust arrows
            % Rotor 1 (front)
            \filldraw[fill=gray!30] (0,2.5,0) ellipse (0.6 and 0.3);
            \node at (0,2.5,0) {1};
            \draw[thick,->,>=stealth,blue] (0,2.5,0) -- (0,2.5,0.8);
            % \draw[thick,->,>=stealth,black,rotate around={330:(0,2.5,0)}] (0,2.5,0) circle (0.4);
            
            % Rotor 2 (right)
            \filldraw[fill=gray!30] (2.5,0,0) ellipse (0.3 and 0.6);
            \node at (2.5,0,0) {2};
            \draw[thick,->,>=stealth,blue] (2.5,0,0) -- (2.5,0,0.8);
            % \draw[thick,->,>=stealth,black,rotate around={60:(2.5,0,0)}] (2.5,0,0) circle (0.4);
            
            % Rotor 3 (back)
            \filldraw[fill=gray!30] (0,-2.5,0) ellipse (0.6 and 0.3);
            \node at (0,-2.5,0) {3};
            \draw[thick,->,>=stealth,blue] (0,-2.5,0) -- (0,-2.5,0.8);
            % \draw[thick,->,>=stealth,black,rotate around={30:(0,-2.5,0)}] (0,-2.5,0) circle (0.4);
            
            % Rotor 4 (left)
            \filldraw[fill=gray!30] (-2.5,0,0) ellipse (0.3 and 0.6);
            \node at (-2.5,0,0) {4};
            \draw[thick,->,>=stealth,blue] (-2.5,0,0) -- (-2.5,0,0.8);
            % \draw[thick,->,>=stealth,black,rotate around={-60:(-2.5,0,0)}] (-2.5,0,0) circle (0.4);
            
            % Central hub
            \filldraw[fill=gray!40] (0,0,0) circle (0.5);
            
            % Body frame - clearly attached to drone center
            \draw[-stealth,red,very thick] (B) -- ++(2,0,0) node[anchor=west]{$x_B$};
            \draw[-stealth,green!70!black,very thick] (B) -- ++(0,2,0) node[anchor=south]{$y_B$};
            \draw[-stealth,blue,very thick] (B) -- ++(0,0,2) node[anchor=south]{$z_B$};
        \end{scope}
    \end{tikzpicture}
    \caption{Quadrotor reference frames and rotor configuration. The world frame axes ($x_W, y_W, z_W$) are fixed, with $z_W$ pointing upward. The body frame ($x_B, y_B, z_B$) is attached to the drone's center of mass with $x_B$ pointing forward. Blue arrows indicate thrust direction, while curved arrows show rotor rotation direction.}
    \label{fig:frames}
\end{figure}
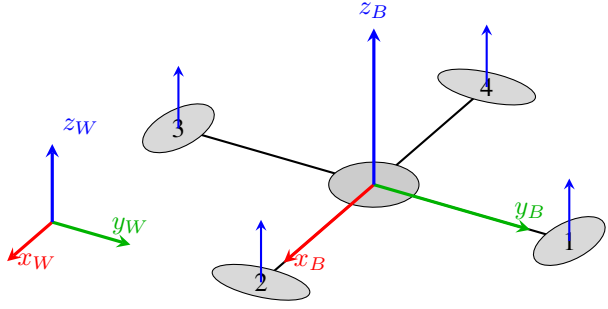

We define a world frame $W$ ($+z_W$ up) and a body frame $B$ ($+x_B$ forward), illustrated in Fig.~\ref{fig:frames}. We use standard notation for position $\mathbf{p}_{WB}$, velocity $\mathbf{v}_{WB}$, Euler angles $\boldsymbol{\Theta} = [\phi, \theta, \psi]^T$, body angular velocity $\boldsymbol{\omega}_B = [p, q, r]^T$, mass $m$, gravity $\mathbf{g}_W$, inertia matrix $\mathbf{J}$, collective thrust $T$ (acting along $-z_B$, so $\mathbf{T}_B = [0 \; 0 \; T]^T$), and body torque $\boldsymbol{\tau}_B$.
\begin{align}
\dot{\mathbf{p}}_{WB} &= \mathbf{v}_{WB} \\
\dot{\boldsymbol{\Theta}} &= \mathbf{W}_{\Theta}(\boldsymbol{\Theta}) \boldsymbol{\omega}_B \label{eq:euler_kinematics_short} \\
\dot{\mathbf{v}}_{WB} &= \mathbf{g}_W + \frac{1}{m}\mathbf{R}_{WB}(\boldsymbol{\Theta}) \mathbf{T}_B \label{eq:linear_dynamics_short} \\
\dot{\boldsymbol{\omega}}_B &= \mathbf{J}^{-1}(\boldsymbol{\tau}_B - \boldsymbol{\omega}_B \times (\mathbf{J} \boldsymbol{\omega}_B)) \label{eq:angular_dynamics_short}
\end{align}
where $\mathbf{R}_{WB}(\boldsymbol{\Theta})$ is the body-to-world rotation matrix and $\mathbf{W}_{\Theta}(\boldsymbol{\Theta})$ maps body rates to Euler rates.

\subsubsection{Hierarchical Controller}
\label{sec:controller_design_revised}
A hierarchical controller translates the reference velocity $\mathbf{v}_{\text{ref}}$ (from policy action via Eq.~\ref{eq:v_desired_rev}) to motor commands (thrust $T$, torques $\boldsymbol{\tau}_B$).

\paragraph{Velocity Tracking Controller (Outer Loop)}
This loop computes desired acceleration $\mathbf{a}_{\text{des}} \in \mathbb{R}^3$ from velocity error $\mathbf{e}_v = \mathbf{v}_{\text{ref}} - \mathbf{v}_{WB}$ via a PD law:
\begin{equation}
\mathbf{a}_{\text{des}} = \mathbf{K}_v \mathbf{e}_v + \mathbf{K}_{vd} \dot{\mathbf{e}}_v
\label{eq:pd_velocity_rev}
\end{equation}
where $\mathbf{K}_v, \mathbf{K}_{vd}$ are diagonal gains; $\dot{\mathbf{e}}_v$ is approximated using finite differences.

\paragraph{Low-Level Controller (Inner Loop)}
The inner loop uses $\mathbf{a}_{\text{des}}$ to compute an intermediate commanded acceleration $\mathbf{a}_{\text{cmd}}$:
\begin{equation}
\mathbf{a}_{\text{cmd}} = \mathbf{K}_{p,\text{pos}} \mathbf{a}_{\text{des}} - \mathbf{K}_{d,\text{pos}} \mathbf{v}_{WB}
\label{eq:acc_cmd_intermediate}
\end{equation}
where $\mathbf{K}_{p,\text{pos}}, \mathbf{K}_{d,\text{pos}}$ are diagonal gains. Collective thrust $T$ is then derived from $\mathbf{a}_{\text{cmd}}$'s vertical component, compensating for gravity:
\begin{equation}
T = m g + m (\mathbf{a}_{\text{cmd}})_z
\label{eq:thrust_calc_code_match}
\end{equation}
Desired roll $\phi_{\text{des}}$ and pitch $\theta_{\text{des}}$ are computed from $\mathbf{a}_{\text{cmd}}$'s lateral components using a simplified mapping:
\begin{equation}
\phi_{\text{des}} = - (\mathbf{a}_{\text{cmd}})_y / g \quad , \quad \theta_{\text{des}} = (\mathbf{a}_{\text{cmd}})_x / g
\label{eq:attitude_des_code_match}
\end{equation}
With desired yaw $\psi_{\text{des}}=0$, the desired attitude $\boldsymbol{\Theta}_{\text{des}} = [\phi_{\text{des}}, \theta_{\text{des}}, \psi_{\text{des}}]^T$ is formed. Body torques $\boldsymbol{\tau}_B$ are then computed by a PD attitude controller:
\begin{equation}
\boldsymbol{\tau}_B = \mathbf{K}_{p,att}(\boldsymbol{\Theta}_{\text{des}} - \boldsymbol{\Theta}) - \mathbf{K}_{d,att}\boldsymbol{\omega}_B
\label{eq:pd_attitude_final}
\end{equation}
where $\boldsymbol{\Theta}, \boldsymbol{\omega}_B$ are current attitude and body rates, and $\mathbf{K}_{p,att}, \mathbf{K}_{d,att}$ are gains.

All specific PD gains (e.g., $\mathbf{K}_v, \mathbf{K}_{p,att}$) were tuned for stability and are provided in our code repository.

\subsection{Racing Agent Design and Training}

This subsection details our core contribution: the CRUISE training framework integrating Proximal Policy Optimization (PPO)~\cite{schulman2017proximal} with curriculum learning and iterative self-play.

\subsubsection{State and Action Space}
\paragraph{State Space}
Each drone agent $i$ observes a comprehensive state vector $\mathbf{o}_{t,i}$ providing proprioceptive and exteroceptive information:
\begin{align}
\mathbf{o}_{t,i} = [& \mathbf{p}_{\text{rel},i}^T, d_{\text{norm},i}, \mathbf{v}_{\text{norm},i}^T, v_{\text{proj},i}, \mathbf{h}_i^T, \mathbf{p}_{\text{norm},i}^T, \nonumber \\
& \sin(\Delta\psi_i), \cos(\Delta\psi_i), \mathbf{P}_{\text{opp},i}^T, \mathbf{D}_{\text{opp},i}^T]^T \label{eq:state_vector}
\end{align}
The components (detailed in our code repository) include: normalized relative position ($\mathbf{p}_{\text{rel},i}$) and distance ($d_{\text{norm},i}$) to the target gate; normalized velocity ($\mathbf{v}_{\text{norm},i}$); projected velocity ($v_{\text{proj},i}$); one-hot target gate index ($\mathbf{h}_i$); normalized world position ($\mathbf{p}_{\text{norm},i}$); yaw difference to gate orientation ($\sin(\Delta\psi_i), \cos(\Delta\psi_i)$); and relative positions/distances to opponents ($\mathbf{P}_{\text{opp},i}, \mathbf{D}_{\text{opp},i}$). Normalization uses environment-dependent factors ($d_{\text{max}}, v_{\text{max}}$).

\paragraph{Action Space}
The policy outputs a continuous, normalized desired acceleration $\mathbf{a}^{\text{raw}}_t \in [-1, 1]^3$. This is processed by clipping to $[-1, 1]$, scaling by a curriculum-dependent agility coefficient $\alpha^k$, and integrating using timestep $\Delta t$ to produce the reference velocity $\mathbf{v}^{\text{desired}}_{t+1}$ for the low-level controller (Eq.~\ref{eq:pd_velocity_rev}):
\begin{equation}
\mathbf{v}^{\text{desired}}_{t+1} = \mathbf{v}_{WB,t} + (\alpha^k \cdot \text{clip}(\mathbf{a}^{\text{raw}}_t, -1, 1)) \cdot \Delta t \label{eq:v_desired_rev}
\end{equation}
This allows the policy to focus on high-level maneuvering commands.

\subsubsection{Reward Function}
\label{sec:reward_rev}

The reward function $\mathcal{R}_k(s_t, a_t)$ shapes agent behavior during curriculum stage $k$ by combining multiple objectives, summarized in Table~\{\ref{tab:reward_summary_rev}\}.

\begin{table}[H] 
\centering
\caption{Summary of Reward Function Component Objectives}
\label{tab:reward_summary_rev}
\footnotesize
\setlength{\tabcolsep}{4pt} % Adjust column separation if needed
\begin{tabular}{@{}lp{0.6\linewidth}@{}} % Adjust width if needed
\hline
\textbf{Component} & \textbf{Objective / Purpose} \\
\hline
$R_{\text{prox}}$ & Encourage proximity to the target gate. \\
$R_{\text{prog}}$ & Reward incremental progress towards the target. \\
$R_{\text{align}}$ & Penalize misalignment with the direct path to the target. \\
$R_{\text{speed}}$ & Encourage speed near the curriculum target $v_{\min}^k$. \\
$R_{\text{over}}$ & Bonus for overtaking another drone. \\
$R_{\text{coll}}$ & Penalty for colliding with other drones. \\
\hline
\end{tabular}
\end{table}

The total reward is a weighted sum:
\begin{equation}
\begin{split}
\mathcal{R}_k(s_t, a_t) = {}& w_{prox} R_{\text{prox}} + w_{prog} R_{\text{prog}} - w_{\text{align}} R_{\text{align}} \\ 
& + w_{speed} R_{\text{speed}} + w_{\text{over}} R_{\text{over}} - w_{\text{coll}} R_{\text{coll}}
\end{split}
\label{eq:reward_total_rev}
\end{equation}
Specific weights ($w_{prox}, \dots, w_{coll}$) and reward component parameters (e.g., $a$ in Eq.~\ref{eq:reward_dist}, $\beta$ in Eq.~\ref{eq:reward_delta}) are curriculum-dependent coefficients, detailed in our code repository. The individual components are defined below. In these equations, $\mathbf{v}_t$ represents the drone's current velocity vector, and $\mathbf{u}_t$ is the unit vector pointing from the drone's position to the center of the target gate. The term $\mathbb{I}(\cdot)$ is the indicator function:

\begin{equation}
R_{\text{prox}}(s_t) = 2 \cdot \frac{e^{-a \cdot x_t} - e^{-a}}{1 - e^{-a}} - 1 
\label{eq:reward_dist}
\end{equation}

\begin{equation}
R_{\text{prog}}(s_t) = \beta \cdot (d_{t-1} - d_t) 
\label{eq:reward_delta}
\end{equation}

\begin{equation}
R_{\text{align}}(s_t) = 
\begin{cases} 
    1 - \dfrac{\mathbf{v}_t \cdot \mathbf{u}_t}{\|\mathbf{v}_t\| \|\mathbf{u}_t\|} & \text{if } \|\mathbf{v}_t\| > \epsilon \\ 
    0 & \text{otherwise}
\end{cases}
\label{eq:reward_dev}
\end{equation}

\begin{equation}
R_{\text{speed}}(s_t) =
\begin{cases} 
v_t - v_{\min}^k, & \text{if } v_t \leq v_{\min}^k \\ 
v_{\min}^k - v_t, & \text{if } v_t > v_{\min}^k 
\end{cases}
\label{eq:reward_speed}
\end{equation}

For $R_{\text{over}}$, a bonus is awarded when agent $i$ overtakes agent $j$. Overtaking is detected when the projection $\text{proj}^t_{ij} = (\mathbf{x}_j - \mathbf{x}_i) \cdot \hat{\mathbf{v}}_i$ (of $j$'s relative position from $i$, onto $i$'s velocity unit vector $\hat{\mathbf{v}}_i$) changes from negative at $t-1$ to positive at $t$:
\begin{equation}
R_{\text{over},i}(s_t) = \sum_{j \neq i} r_{\text{over}} \cdot \mathbb{I}\left[(\text{proj}^{t-1}_{ij} < 0) \wedge (\text{proj}^{t}_{ij} > 0)\right] 
\label{eq:reward_over}
\end{equation}

\begin{equation}
R_{\text{coll},i}(s_t) = \mathbb{I}( \exists j \neq i : \|\mathbf{p}_{WB,i} - \mathbf{p}_{WB,j}\| < \delta ) 
\label{eq:reward_collision}
\end{equation}

\subsubsection{Curriculum Learning Implementation}
\label{sec:curriculum_short}

To facilitate efficient and robust policy learning, we employ a \textbf{five-stage curriculum} ($k = 1, \dots, 5$), where task difficulty and realism are systematically increased. Each stage $k$ is defined by environment and reward parameters $\Theta_k$:
\begin{equation}
\Theta_k = \left\{ v_{\text{min}}^k, \alpha^k, c_{\text{enable}}^k, w_{\text{coll}}^k, g_{\text{tol}}^k, w_{\text{over}}^k \right\}
\end{equation}
where $v_{\text{min}}^k$ is the minimum reference velocity target (m/s), $\alpha^k$ action scaling (agility coefficient), $c_{\text{enable}}^k$ indicates if collision penalties are active, $w_{\text{coll}}^k$ the collision penalty magnitude, $g_{\text{tol}}^k$ allowable gate passage tolerance (m), and $w_{\text{over}}^k$ the reward for overtaking. Parameter values for each stage are detailed in Table~\ref{tab:curriculum_short}.

\begin{table}[H]
\centering
\caption{Curriculum Learning Stage Parameters (Qualitative Trends)} 
\label{tab:curriculum_short}
\footnotesize
\setlength{\tabcolsep}{4pt}
\begin{tabular}{@{}ccccccccc@{}}
\hline
\textbf{Stage} & \textbf{Name} & \textbf{Timesteps} & $v_{\text{min}}^k$ & $\alpha^k$ & $c_{\text{enable}}^k$ & $w_{\text{coll}}^k$ & $g_{\text{tol}}^k$ & $w_{\text{over}}^k$ \\
\hline
1 & Basics & $1 \times 10^6$ & 1.0 & 2.0 & False & 0.0 & 0.5 & 0.0 \\
2 & Intermediate & $3 \times 10^6$ & 3.0 & 3.0 & True & 0.25 & 0.3 & 0.1 \\
3 & Advanced & $6 \times 10^6$ & 5.0 & 4.0 & True & 0.5 & 0.25 & 0.2 \\
4 & Advanced II & $1 \times 10^7$ & 7.0 & 6.0 & True* & 0.6 & 0.2 & 0.2 \\
5 & Advanced III & $2 \times 10^7$ & 10.0 & 7.5 & True* & 0.7 & 0.2 & 0.2 \\
\hline
\end{tabular}
\begin{flushleft}
\footnotesize
\textbf{Note:} *From Stage 4 onward, collisions not only incur a penalty but also terminate the episode. 
\end{flushleft}
\end{table}

\subparagraph{Curriculum Design and Progression}
\label{par:curriculum_rationale}

Our curriculum is designed to progressively build the necessary skills for high-speed racing by guiding the agent through five distinct stages (see Table~\ref{tab:curriculum_short} for specific parameter values). The stages are as follows:
\begin{itemize}
\item \textbf{Stage 1 (Basics):} Focuses on fundamental navigation. By using a low speed and disabling collision penalties, this stage allows the agent to learn the core task of flying through gates without being prematurely punished for early mistakes, thus solving the initial exploration problem.
\item \textbf{Stage 2 (Intermediate):} Introduces penalties for colliding with static obstacles (i.e., the track walls) while increasing speed. This stage begins to shape the agent's policy to balance speed with safety and precision.
\item \textbf{Stage 3 (Advanced):} Further increases speed and penalties, refining the agent's ability to fly aggressively while respecting the track boundaries.
\item \textbf{Stage 4 (Advanced II):} Critically, collisions are now configured to be \textbf{terminal events}. This strict condition forces the agent to develop a truly robust policy, as even a single error ends the attempt.
\item \textbf{Stage 5 (Advanced III):} Pushes the agent to its limits with maximum speed and the tightest tolerances. This stage polishes the policy, ensuring it is capable of handling the most extreme conditions before facing dynamic opponents.
\end{itemize}
To ensure a smooth adaptation as difficulty increases, policy weights are transferred after each stage ($\theta_{k+1}^{\text{init}} = \theta_k$), allowing the agent to build directly on prior competencies. The progression incrementally increases difficulty towards the system's dynamic limits (e.g., $v_{\text{min}}^5 = 10.0\,\text{m/s}$), balancing safe training with high-performance evaluation. Upon completion of this single-agent curriculum, the trained policy is then used to initialize the iterative self-play phase for multi-agent racing, where agents learn to handle the complexity of dynamic opponents.

\subsubsection{Self-Play Methodology}
\label{sec:self_play_rev}

To train competitive policies efficiently and scalably ($O(n)$ complexity), we employ an \textit{iterative self-play} mechanism, detailed in Algorithm~\ref{alg:self_play_rev}.

\subparagraph{Training and Opponent Update}
A single active policy $\pi_{\theta_a}$ is trained via PPO against $n-1$ frozen opponent policies $\{\pi_{\phi_i}\}$. These opponents are periodically updated by copying the active policy's weights ($\phi_i \leftarrow \theta_a$ for all $i$) if $\pi_{\theta_a}$'s win rate $w$ against them exceeds a predefined threshold $\tau$. The win rate $w$ is computed over $n_{\text{eval}}$ evaluation episodes as:
\begin{equation}
w = \frac{1}{n_{\text{eval}}} \sum_{j=1}^{n_{\text{eval}}} \mathbb{I}\left( p_a^{(j)} > \max_{k} p_k^{(j)} \right)
\label{eq:opponent_update_cond}
\end{equation}
where $p_a^{(j)}$ and $p_k^{(j)}$ are progress metrics (gates passed) for the active agent and opponent $k$ respectively in evaluation episode $j$, and $\mathbb{I}(\cdot)$ is the indicator function. This mechanism ensures that $\pi_{\theta_a}$ continually strives against increasingly competent versions of itself. Specific evaluation frequencies ($T_{\text{eval}}$ in Alg.~\ref{alg:self_play_rev}) and the win rate threshold ($\tau$) are hyperparameters detailed in our code repository.

\begin{algorithm}[H]
\caption{Iterative Self-Play Training within Curriculum}
\label{alg:self_play_rev}
\begin{algorithmic}[1]
\State Initialize active policy $\pi_{\theta_a}$ and opponent policies $\{\pi_{\phi_i}\}_{i=1}^{n-1}$ with identical parameters
\State $\phi_i \leftarrow \theta_a$ for $i \in \{1, 2, ..., n-1\}$ \Comment{Initial sync}
\For{each curriculum stage $k = 1$ to $K$}
    \State Set environment parameters $\Theta_k$ from Table~\ref{tab:curriculum_short}
    \For{timestep $t = 1$ to $N_k$} \Comment{$N_k$: budget per stage}
        \State Collect experience tuple $(s_t, a_t, r_t, s_{t+1})$ using $\pi_{\theta_a}$ against frozen $\{\pi_{\phi_i}\}$
        \State Update $\pi_{\theta_a}$ using PPO and collected experience
        \If{$t \pmod{T_{\text{eval}}} = 0$} \Comment{Periodic evaluation}
            \State Evaluate $\pi_{\theta_a}$ against current $\{\pi_{\phi_i}\}$ over $M$ episodes
            \State Calculate win rate $w$ based on $p_a$ (Eq.~\ref{eq:opponent_update_cond})
            \If{$w \geq \tau$} \Comment{Check mastery threshold}
                \State Update opponent policies: $\phi_i \leftarrow \theta_a$ for all $i$ \Comment{Sync opponents}
            \EndIf
        \EndIf
    \EndFor
\EndFor
\end{algorithmic}
\end{algorithm}

\subparagraph{Self-Play Dynamics}
This iterative process inherently handles the non-stationarity of evolving opponents. Self-play, operating within each curriculum stage, creates a synergistic learning dynamic: agents adapt to increasingly harder tasks (via curriculum) while simultaneously improving against more competent opponents (via self-play). This is crucial for developing robust and competitive behaviors.

\section{Results and Discussion}
\label{sec:results}

This section empirically evaluates CRUISE, detailing the experimental setup, metrics, baselines, quantitative results, and an ablation study on the curriculum's impact.

\subsection{Experimental Setup}
\label{sec:setup}

\begin{figure}[t]
\centering
\includegraphics[width=\linewidth]{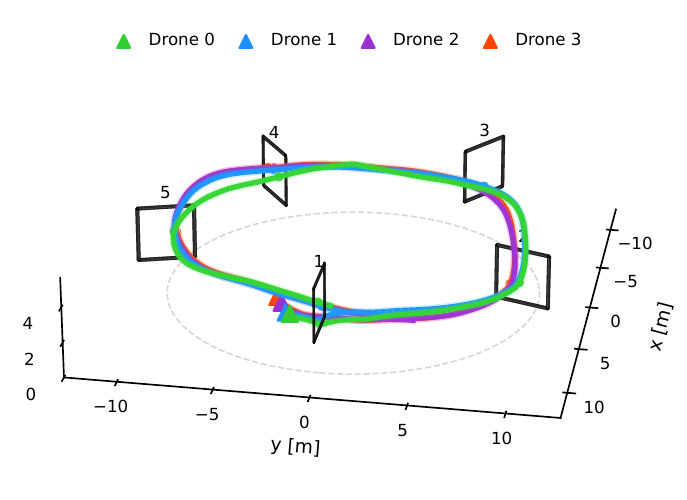}
\caption{Representative trajectories of four CRUISE-trained drones navigating the Ring Track. The track features five gates (black rectangles) in a circular layout with alternating heights. Distinct colors show individual drone paths, illustrating coordinated high-speed navigation.}
\label{fig:ring_track}
\end{figure}

\begin{figure}[t]
\centering
\includegraphics[width=\linewidth]{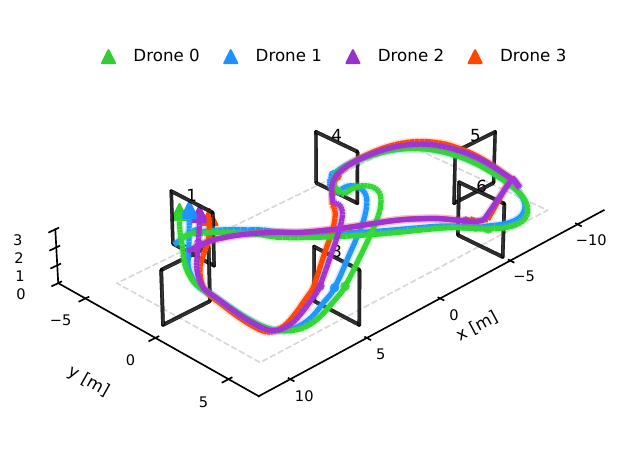} % Replace with your path
\caption{Representative trajectories of four CRUISE-trained drones on the more complex Figure-Eight Track. The six gates form intersecting loops, requiring intricate coordination at crossing points.}
\label{fig:figure_eight_track}
\end{figure}

CRUISE was evaluated with $N=\{2, 3, 4\}$ drones in a custom Gymnasium multi-agent environment with realistic quadrotor dynamics (Sec.~\ref{sec:methodology}). Two tracks tested different capabilities: a \textbf{Ring Track} for basic 3D maneuvering and speed (Fig.~\ref{fig:ring_track}), and a more complex \textbf{Figure-Eight Track} with intersections for coordination (Fig.~\ref{fig:figure_eight_track}). Each configuration ran for 100 episodes with randomized starts for robustness.

\subsection{Performance Metrics}
\label{sec:metrics}

Performance was quantified by three primary metrics (averaged over successful episodes): \textbf{Mean Velocity} (\si{m/s}) indicating agility, \textbf{Lap Completion Time} (\si{s}), and \textbf{Success Rate} (\si{\percent}) measuring collision-free completion.

\subsection{Baseline Methods}
\label{sec:baselines}

CRUISE is compared against two baselines:
\begin{itemize}
    \item \textbf{SE-IBR:} 
    To benchmark our framework against the state of the art in competitive MARL, we compare it against Sensitivity Enhanced Iterative Best Responsee (SE-IBR) \cite{wang2020multi}. While methods like MAPPO are standard for cooperative tasks, our zero-sum drone racing environment is fundamentally competitive. SE-IBR, a game-theoretic planner, is therefore a more theoretically appropriate and challenging benchmark. As no public code was available, we re-implemented the method based on its described iterative best response (IBR) mechanism and opponent trajectory inference for 3D racing. 
    \item \textbf{VANILLA:}  To isolate and  validate the contribution of our curriculum, we use VANILLA baseline. This agent uses the same PPO algorithm and final reward function as CRUISE, but is trained from scratch only on the final Stage 5 task (see Table \ref{tab:curriculum_short}), without any of the intermediate curriculum stages. This baseline therefore serves as a direct and intuitive ablation of the curriculum structure, allowing us to measure the performance of an agent that does not benefit from the guided learning sequence.
\end{itemize}

\subsection{Quantitative Evaluation and Comparison}
\label{sec:evaluation}

Tables~\ref{tab:ring_track_results} and~\ref{tab:figure_eight_results} present the quantitative comparison, showing CRUISE's superior performance across both tracks and drone densities.

On the Ring Track (Table~\ref{tab:ring_track_results}), CRUISE achieved superior mean velocities (\SIrange{4.3}{4.4}{m/s}) and lap times (\SIrange{13.3}{13.6}{s}) over the slower SE-IBR (\SI{\approx 2.3}{m/s}) and VANILLA (\SI{2.8}{m/s} for N=2). CRUISE maintained high success rates (\SIrange{91}{100}{\percent}), while SE-IBR's safety dropped at N=4 (\SI{50}{\percent}) and VANILLA failed for $N \ge 3$.

On the more challenging Figure-Eight Track (Table~\ref{tab:figure_eight_results}), CRUISE again demonstrated superior speed (mean velocity \SIrange{3.4}{3.5}{m/s}, lap times \SIrange{16.1}{16.7}{s}) over SE-IBR (\SIrange{1.6}{1.8}{m/s}) and VANILLA (\SIrange{1.8}{2.1}{m/s}). CRUISE also achieved high, consistent success rates (\SIrange{97}{98}{\percent}), while VANILLA's dropped sharply with density. SE-IBR maintained high safety (\SIrange{90}{100}{\percent}) but at significantly lower speeds than CRUISE. These results show CRUISE robustly balances speed and safety in complex coordination scenarios.

\begin{table}[htbp]
\centering
\caption{Performance comparison on the Ring Track. Values are mean $\pm$ std dev over 100 episodes.}
\label{tab:ring_track_results}
\footnotesize
\renewcommand{\arraystretch}{1.0}
\begin{tabular}{lccc}
\hline
\multirow{2}{*}{\textbf{Method}} & \textbf{Lap Time} & \textbf{Velocity} & \textbf{Success} \\
 & (s) & (m/s) & Rate (\%) \\
\hline
\multicolumn{4}{l}{\textit{\textbf{2 drones}}} \\
CRUISE & \textbf{13.3 $\pm$ 0.1} & \textbf{4.4 $\pm$ 0.0} & \textbf{100} \\
SE-IBR & 26.0 $\pm$ 1.0 & 2.3 $\pm$ 0.1 & \textbf{100} \\
VANILLA & 20.9 $\pm$ 0.0 & 2.8 $\pm$ 0.0 & 50 \\
\hline
\multicolumn{4}{l}{\textit{\textbf{3 drones}}} \\
CRUISE & \textbf{13.4 $\pm$ 0.2} & \textbf{4.4 $\pm$ 0.1} & \textbf{91.3} \\
SE-IBR & 26.2 $\pm$ 0.7 & 2.2 $\pm$ 0.1 & 90 \\
VANILLA & N/A  & N/A  & 0 \\
\hline
\multicolumn{4}{l}{\textit{\textbf{4 drones}}} \\
CRUISE & \textbf{13.6 $\pm$ 0.5} & \textbf{4.3 $\pm$ 0.2} & \textbf{93.5} \\
SE-IBR & 26.1 $\pm$ 0.8 & 2.3 $\pm$ 0.1 & 50 \\
VANILLA & N/A  & N/A & 0 \\
\hline
\end{tabular}
\end{table}

\begin{table}[htbp]
\centering
\caption{Performance comparison on the Figure-Eight Track. Values are mean $\pm$ std dev over 100 episodes.}
\label{tab:figure_eight_results}
\footnotesize
\renewcommand{\arraystretch}{1.0}
\begin{tabular}{lccc}
\hline
\multirow{2}{*}{\textbf{Method}} & \textbf{Lap Time} & \textbf{Velocity} & \textbf{Success} \\
 & (s) & (m/s) & Rate (\%) \\
\hline
\multicolumn{4}{l}{\textit{\textbf{2 drones}}} \\
CRUISE & \textbf{16.1 $\pm$ 0.3} & \textbf{3.5 $\pm$ 0.1} & 98 \\
SE-IBR & 30.9 $\pm$ 0.8 & 1.8 $\pm$ 0.1 & 90 \\
VANILLA & 27.4 $\pm$ 0.1 & 2.1 $\pm$ 0.0 & \textbf{100} \\
\hline
\multicolumn{4}{l}{\textit{\textbf{3 drones}}} \\
CRUISE & \textbf{16.2 $\pm$ 1.7} & \textbf{3.5 $\pm$ 0.4} & \textbf{98} \\
SE-IBR & 34.0 $\pm$ 3.7 & 1.7 $\pm$ 0.2 & 90 \\
VANILLA & 28.3 $\pm$ 0.2 & 2.0 $\pm$ 0.0 & 43.3 \\
\hline
\multicolumn{4}{l}{\textit{\textbf{4 drones}}} \\
CRUISE & \textbf{16.7 $\pm$ 0.7} & \textbf{3.4 $\pm$ 0.1} & 97 \\
SE-IBR & 36.6 $\pm$ 6.4 & 1.6 $\pm$ 0.3 & \textbf{100} \\
VANILLA & 31.1 $\pm$ 0.9 & 1.8 $\pm$ 0.1 & 30 \\
\hline
\end{tabular}
\end{table}

\subsection{Ablation Study: Impact of Curriculum Learning}
\label{sec:ablation}

\begin{figure}[t]
\centering
\includegraphics[width=\linewidth]{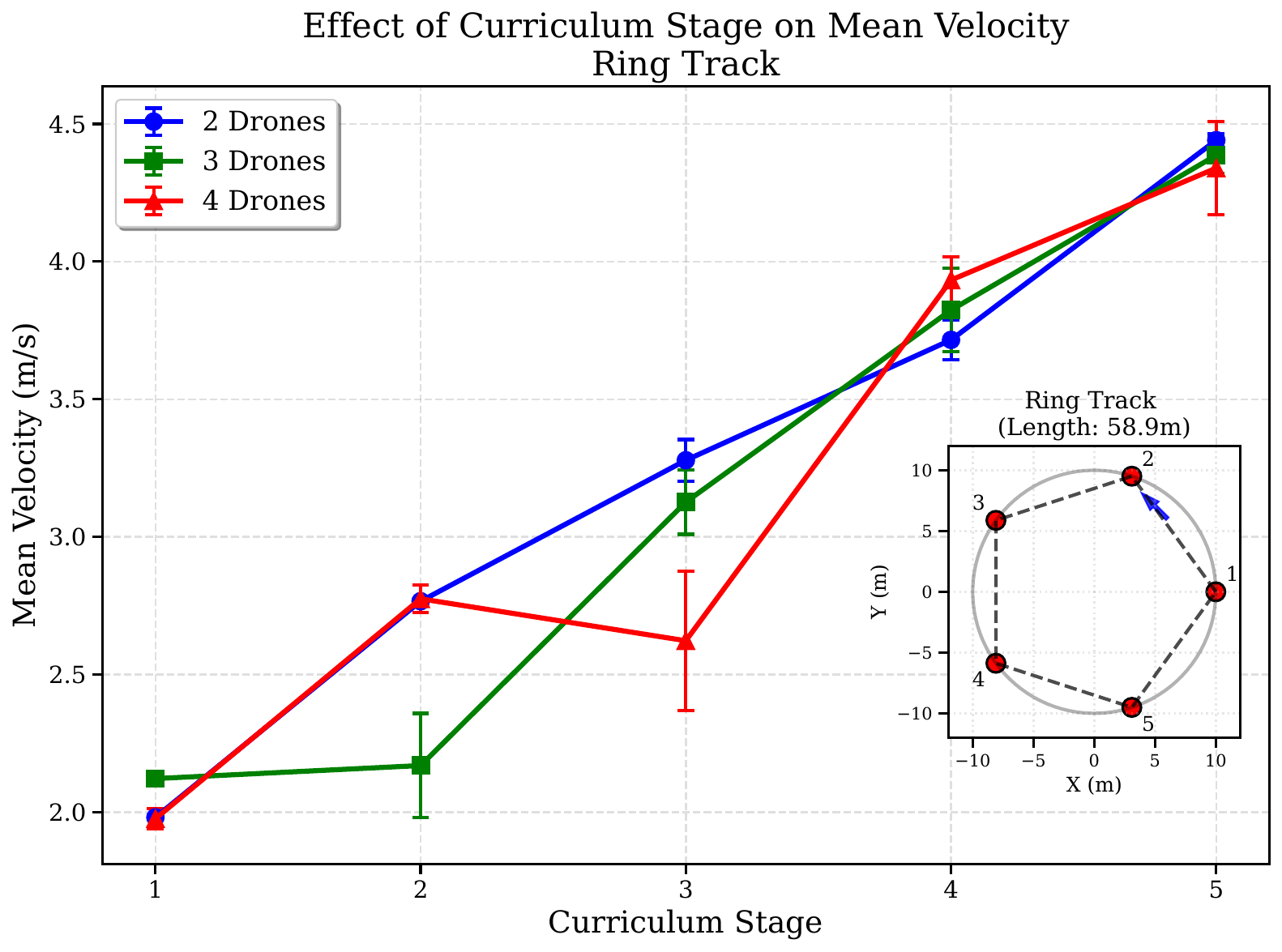} % Replace with path
\caption{Ablation study on the Ring Track: Mean velocity (\si{m/s}) vs. curriculum stage for $N=2, 3, 4$ drones. Error bars indicate std dev (100 trials). Inset shows track layout. Increasing velocity across stages demonstrates the curriculum's benefit.}
\label{fig:ablation_ring}
\end{figure}

\begin{figure}[t]
\centering
\includegraphics[width=\linewidth]{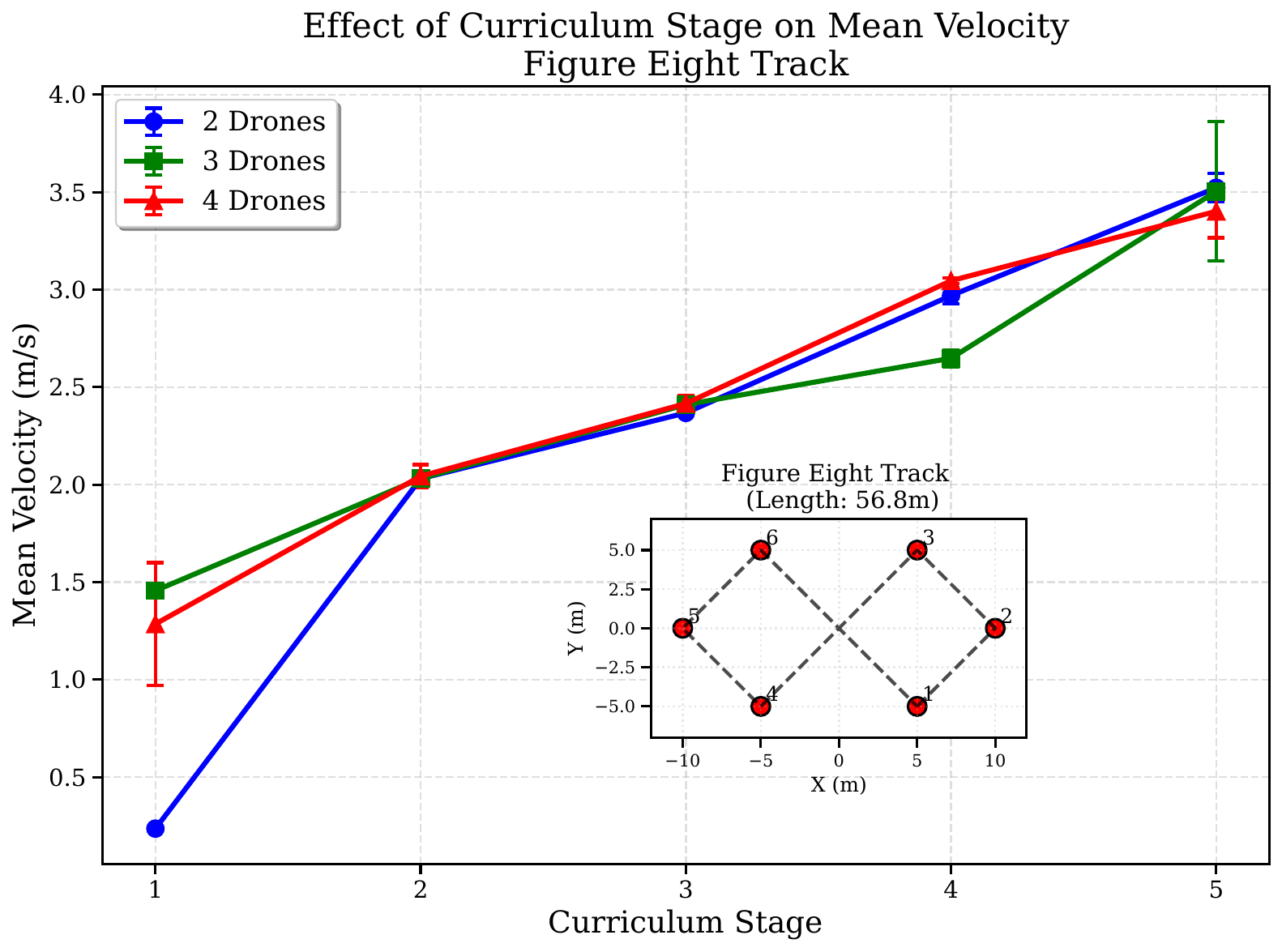} % Replace with path
\caption{Ablation study on the Figure-Eight Track: Mean velocity (\si{m/s}) vs. curriculum stage for $N=2, 3, 4$ drones. Error bars show std dev (100 trials). Inset shows track layout. Performance increases consistently with curriculum stage.}
\label{fig:ablation_eight}
\end{figure}

The curriculum's contribution was assessed by measuring mean velocity after each training stage. For all tracks and drone counts ($N=2, 3, 4$), velocity consistently increased with curriculum progression (Figs.~\ref{fig:ablation_ring} and ~\ref{fig:ablation_eight}), confirming that the curriculum effectively scaffolds skill acquisition for high-speed flight. The VANILLA baseline's poor performance (Tables~\ref{tab:ring_track_results} and~\ref{tab:figure_eight_results}) further highlights the curriculum's necessity.

\subsection{Discussion}
\label{sec:discussion}

The experimental results validate CRUISE's core hypotheses: that a structured curriculum is essential for sample efficiency and that this foundation enables scalable learning of competitive strategies through self-play.

The curriculum is the key to solving the exploration problem. This is most clearly demonstrated by the VANILLA baseline, which serves as a direct ablation of the curriculum. Its complete failure to converge in scenarios with three or more agents (Tables~\ref{tab:ring_track_results}, \ref{tab:figure_eight_results}) underscores the severity of the exploration challenge in this domain. By breaking the problem down into manageable stages, CRUISE ensures the agent first masters fundamental flight control before tackling the complexities of high-speed, multi-agent interaction. This guided process makes efficient use of the sample budget, leading to continuous and effective learning where a direct approach fails.

CRUISE enables the emergence of robust, non-greedy strategies. A crucial insight is that the final policies do not simply maximize velocity. Instead, agents converge to effective racing speeds (e.g., \SI{\approx 4.4}{m/s} on Ring) that are well below the maximum reference velocity (\SI{10}{m/s}) available in the final curriculum stage. This demonstrates that the framework successfully balanced competing reward terms, fostering a sophisticated policy that prioritizes consistency and collision avoidance over raw speed, leading to higher success rates.

The synergy between the curriculum and self-play stabilizes multi-agent learning. Our simplified iterative self-play mechanism, which uses frozen opponents, proves highly effective. We posit this is because the single-agent curriculum provides such a strong policy initialization. Agents enter the self-play phase as already-competent pilots, allowing the learning to focus on refining opponent interaction rather than struggling with basic flight. This strong prior mitigates the non-stationarity problem inherent in MARL and allows for a stable, scalable training process. Furthermore, the progressively additive nature of the curriculum, where skills are built upon rather than replaced, inherently mitigates catastrophic forgetting, ensuring foundational competencies are retained and strengthened throughout training.

\subsection{Limitations and Broader Context}
\label{sec:limitations}

While CRUISE demonstrates a robust and scalable training methodology, we identify three key areas for future work.

First, the sim-to-real gap remains the primary hurdle. This work serves as a foundational study in simulation. Deploying these high-speed, interactive policies on physical hardware will require addressing challenges such as unmodeled dynamics, sensor noise, and communication latency. A systematic investigation into domain randomization and robust controller design for sim-to-real transfer is the most critical next step.

Second, the reward function is hand-crafted. While effective, the design of the reward components and their respective weights required significant domain expertise and empirical tuning. This is a common challenge in applying RL to complex robotics tasks. Future work could explore automating this process through techniques like inverse reinforcement learning (IRL) from expert demonstrations or population-based methods that co-evolve reward functions alongside policies.

Finally, the self-play mechanism can be extended. Our iterative approach with a single updating policy is efficient and stable but represents one point in a wide spectrum of self-play strategies. Exploring more advanced schemes, such as training against a diverse league of past policies or using population-based training, could potentially lead to more generalizable and unpredictable strategies, further pushing the boundaries of competitive multi-agent performance.

\section{Conclusion}
\label{sec:conclusion}

This paper presented CRUISE, a framework that successfully addresses the critical challenges of sample efficiency and scalability in competitive multi-drone racing. We demonstrated that a structured curriculum is not merely an enhancement but a necessity for solving the severe exploration problem inherent to this complex domain, a finding validated by our ablation studies. This curriculum provides a crucial foundation of single-agent competency, upon which an efficient iterative self-play mechanism builds emergent, coordinated strategies for crowded environments. The resulting policies significantly outperform standard RL and game-theoretic baselines, exhibiting a robust balance between speed and safety. Ultimately, CRUISE provides a blueprint for structuring the learning process itself to unlock complex, emergent behaviors in challenging multi-agent robotic systems. To facilitate further research, our full source code and pretrained models are publicly available at \url{https://doi.org/10.5281/zenodo.17256943}.

\section{Future Work}
\label{sec:future_work}

Building on this foundation, our future work will proceed along three primary axes. First and foremost is bridging the sim-to-real gap, which involves deploying and validating these policies on physical quadrotors while addressing challenges like dynamics mismatch and sensor noise. Second, we aim to reduce the reliance on hand-crafted rewards by exploring techniques like inverse reinforcement learning to automate the reward engineering process. Finally, we will investigate more advanced self-play schemes, such as training against a diverse league of past policies, to foster even more generalizable and unpredictable competitive strategies.

%Bibliography
% \bibliographystyle{unsrt}  
\bibliographystyle{elsarticle-harv} 
\bibliography{References}

\end{document}